%% file: neurips_2026.tex
\documentclass{article}

\usepackage[preprint]{neurips_2026}

\usepackage[utf8]{inputenc} 
\usepackage[T1]{fontenc}    
\usepackage{hyperref}       
\usepackage{url}            
\usepackage{graphicx}       
\usepackage{booktabs}       
\usepackage{amsfonts}       
\usepackage{nicefrac}       
\usepackage{microtype}      
\usepackage{xcolor}         
\usepackage{colortbl}
\usepackage{multirow}
\usepackage{wrapfig}
\usepackage{amsmath}
\usepackage{caption}

\definecolor{cvprblue}{rgb}{0.21,0.49,0.74}

\newlength\savewidth
\newcommand\shline{\noalign{\global\savewidth\arrayrulewidth
  \global\arrayrulewidth 1pt}\hline\noalign{\global\arrayrulewidth\savewidth}}

\newcommand{\tablestyle}[2]{\setlength{\tabcolsep}{#1}\renewcommand{\arraystretch}{#2}\centering\footnotesize}

\title{RATS! Patches Talk Through Registers:\\ Emergent Parts in Register Attention Transformers}

\author{
  Timing Yang$^{1}$ \quad
  Predrag Neskovic$^{2}$ \quad
  Jansen Seheult$^{3}$ \quad
  Wenchao Han$^{3}$ \\[2pt]
  \bfseries Anand Bhattad$^{1}$ \quad
  Alan Yuille$^{1}$ \quad
  Feng Wang$^{1*}$ \\[6pt]
  $^{1}$Johns Hopkins University 
  $^{2}$Office of Naval Research, Arlington, VA \\
  $^{3}$Department of Laboratory Medicine and Pathology, Mayo Clinic, MN, USA \\[4pt]
}

\begin{document}

\maketitle

\begin{center}
    \vspace{-0.8cm}
    \includegraphics[width=0.95\linewidth]{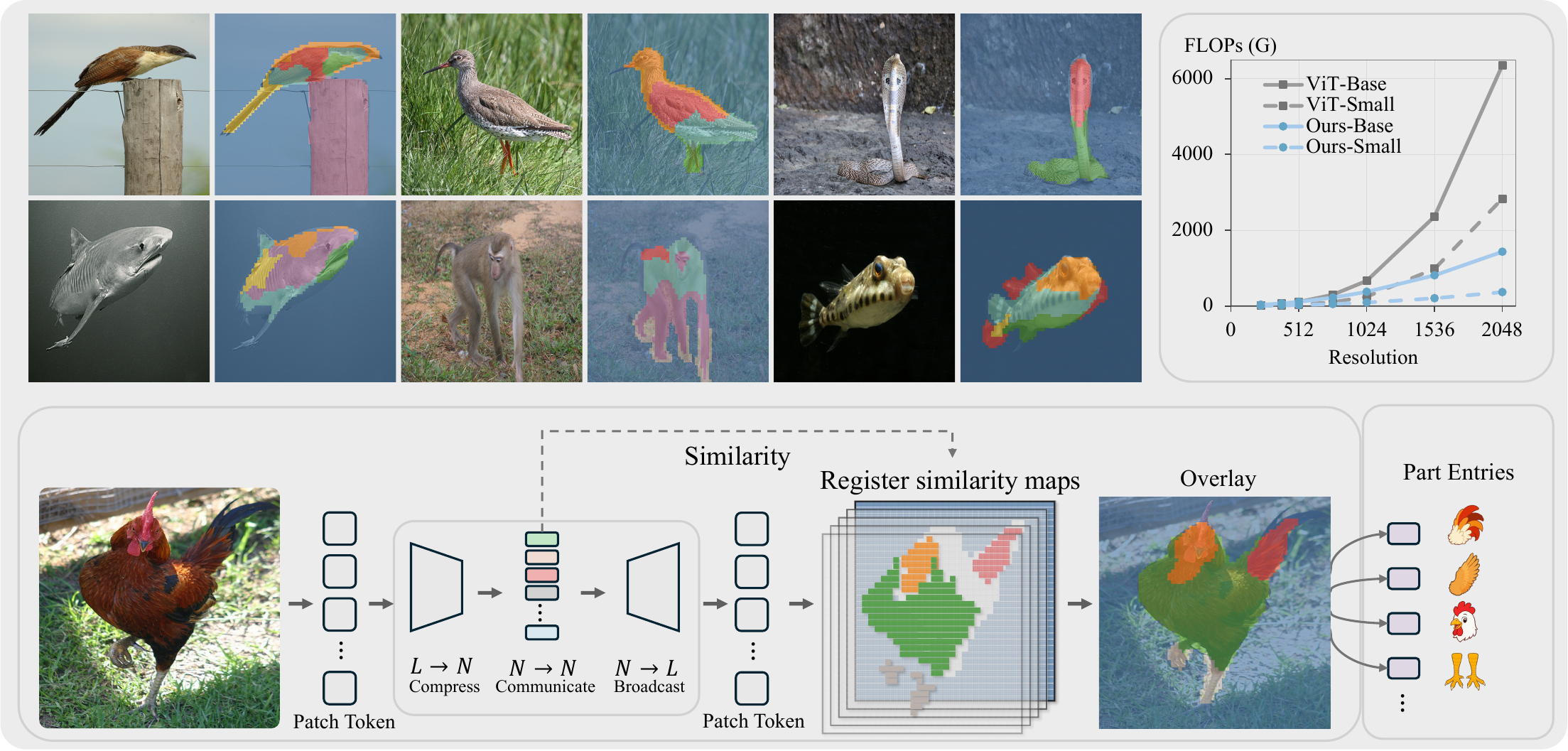}
    \captionof{figure}{
    \textbf{\textit{RATS}} discovers compositional part structure through a register-token bottleneck. \emph{Top:} unsupervised part segmentations emerge across diverse categories (birds, snake, shark, monkey, fish). \emph{Top-right:} the bottleneck makes inference scale  efficiently with image resolution. \emph{Bottom:} each block routes patch--patch communication through $N$ register tokens via a three-step \emph{compress--communicate--broadcast} attention; per-register similarity maps over the patch grid recover semantic parts, which we catalog into a small dictionary of reusable \emph{part entries}.
    }
    
    \label{fig:teaser}
\end{center}

\input{sec/0_abs.tex}
\input{sec/1_intro.tex}
\input{sec/2_related.tex}
\input{sec/3_method.tex}

\input{sec/4_exp.tex}
\input{sec/5_con.tex}

\bibliographystyle{plain}
\bibliography{references}

\newpage
\appendix
\input{sec/6_appendix.tex}



\end{document}

%% file: sec/0_abs.tex
\begin{abstract}

\def\thefootnote{*}\footnotetext{Corresponding author. Email: \url{wangf3014@gmail.com}. Code: \url{https://github.com/yangtiming/RATS}.}

When humans see a bird, they recognize far more than just ``bird'' --- they see a head, wings, and talons, a structured assembly of reusable parts that can be identified across every bird they have ever seen. We ask whether a self-supervised visual model can discover the same compositional structure on its own. To this end, we propose \textbf{RATS} (\textbf{R}egister \textbf{A}ttention \textbf{T}ransformer\textbf{s}), which decomposes the classification token into $N$ learnable register tokens that route patch information through an $L{\to}N{\to}N{\to}L$ bottleneck via a three-step \emph{compress--communicate--broadcast} attention. The $N$ registers are partitioned across the $H$ attention heads, so that registers assigned to different heads do not interact with each other. Without auxiliary losses or part annotations, each register spontaneously specializes into a proto-semantic region whose emerging structure resembles object parts. RATS surpasses all baselines by +12 mIoU on average across five segmentation benchmarks, with consistent gains on ADE20K (+1.11 mIoU) and COCO (+0.2 AP$^{\text{m}}$). Its register dictionary further exhibits part-level consistency and semantic proximity across related categories. Our results suggest that RATS may provide a useful architectural prior for structured and interpretable visual representation learning.
\end{abstract}

%% file: sec/1_intro.tex
\section{Introduction}

When a person looks at a photograph of a bird, they do not merely recognize ``bird.'' They see a head, two wings, talons gripping a branch --- a structured assembly of parts, each meaningful on its own and recognizable across every bird they have ever seen. This decomposition is immediate and effortless: humans naturally parse visual scenes into reusable local structures, whether a knife's metal blade and wooden handle, or a car's tires, body, and windows. These parts are \emph{transferable} across instances (all cats have ears), \emph{shared} across categories (bicycle tires and car tires are fundamentally similar), and \emph{composable} (different parts combine into different objects). They are, in a precise sense, the atomic vocabulary of visual understanding --- and, we argue, the representations that a visual model should discover on its own.

Self-supervised learning has produced surprisingly powerful visual representations.
Methods ranging from masked patch prediction~\cite{he2022masked}
and cross-view consistency~\cite{chen2020simple} to self-distillation~\cite{caron2021emerging}
have shown that rich structure can be learned from images alone, without any labels.
The DINO family in particular has pushed this further: DINOv2~\cite{oquab2024dinov2}
features, when visualized via PCA, reveal striking cross-image correspondences---the same
principal component highlights cat ears, dog ears, and rabbit ears simultaneously, or maps
bicycle wheels onto car wheels across entirely different images. Part-level structure, it turns out,
is already latent in self-supervised features. Yet surfacing it as an explicit, structured
representation remains an open problem. Standard ViTs have no mechanism to do so:
fully-connected self-attention collapses everything into a single global classification token(\texttt{[CLS]}), with no
internal token ever asked to represent a specific region or part. The patch tokens, while
individually informative, are never grouped into coherent regions---the model has no
representational commitment to parts. This suggests a natural next step: rather than treating an
image as a flat bag of patch tokens to be globally summarized, can we \emph{group} these tokens
into meaningful region-level representations---letting the model explicitly organize what it has
implicitly learned?

Vision Transformers provide no explicit grouping mechanism, and several lines of work address this by introducing intermediate latent tokens as grouping containers.
Superpixel-based methods~\cite{mei2024spformer} follow image boundaries but are restricted to local context, lacking the global information needed for part identity.
Object-centric methods~\cite{locatello2020object,seitzer2023dinosaur,fan2024adaslot} let slot vectors compete to represent scene elements, but reconstruction-driven supervision yields object-level masks rather than part-level semantics, and hard mutual exclusion breaks semantic coherence.
GroupViT~\cite{xu2022groupvit} achieves semantic grouping but requires image-text pairs, specializing tokens to text-described categories rather than structural parts.
Perceiver IO~\cite{jaegle2021perceiver} compresses inputs into a compact latent array, but each latent is jointly modeled by all attention heads, leading to fully mixed representations that lack independent semantics.

Building on these observations, we propose \textbf{RATS} (\textbf{R}egister 
\textbf{A}ttention \textbf{T}ransformer\textbf{s}), aiming to let self-supervised 
learning spontaneously surface part-level representations. Concretely, we introduce an $L{\to}N{\to}N{\to}L$ bottleneck in each transformer block that decomposes 
the \texttt{[CLS]} aggregation into $N$ register tokens: patches aggregate into 
registers (\emph{compress}), registers exchange information (\emph{communicate}), 
and broadcast back to patches (\emph{broadcast}). The $N$ registers are partitioned exclusively across the $H$ attention heads, so each head owns an independent register subset that communicates only within the head. Confining each head to its own subspace encourages different heads to specialize to different regions, so part-level structure can emerge (Figure~\ref{fig:teaser}). On five segmentation benchmarks, RATS demonstrates stronger  part-level grouping than all existing baselines ($+12$ mIoU). By providing  register tokens as initial query tokens, RATS further yields  a stronger initialization for dense prediction, outperforming ViT on ADE20K  ($+1.11$ mIoU) and COCO ($\text{AP}^m$ $+0.2$). Finally, the visual  dictionary extracted from the trained registers reveals a level semantic  organization: within-supercategory part consistency and cross-category  taxonomic proximity.

%% file: sec/2_related.tex
\section{Related Work}

\paragraph{Self-Supervised Visual Representation Learning and Segmentation.}
Self-supervised learning has progressed from pretext
tasks~\cite{gidaris2018unsupervised,noroozi2016unsupervised,zhang2016colorful}
through contrastive~\cite{he2020momentum,chen2020simple,caron2020swav} and
non-contrastive~\cite{grill2020bootstrap,chen2021exploring,bardes2022vicreg}
methods, to self-distillation on Vision
Transformers~\cite{caron2021emerging} and masked image
modeling~\cite{he2022masked,bao2022beit,xie2022simmim,zhou2022ibot,wei2022masked}.
The DINO
family~\cite{caron2021emerging,oquab2024dinov2,simeoni2025dinov3} has
demonstrated that self-supervised training can yield patch tokens with
strong part-level features alongside a global CLS token, as revealed by
PCA visualization of the patch
tokens. Several
works~\cite{amir2022deep,melas2022deep,wang2023tokencut,hung2019scops,choudhury2021unsupervised,hamilton2022stego,ziegler2022leopart,aniraj2024pdiscoformer,wang2024sclip}
are related to part/object segmentation and discovery, but these methods fundamentally still extract features from
patch tokens. To the best of our knowledge, no existing method provides
an architectural mechanism to explicitly extract part-level features
through register tokens. RATS addresses this by introducing a register token bottleneck that
extracts part-level information from patch tokens into
register tokens.

\paragraph{Register Tokens in Visual Representation.}
Special-purpose learnable tokens have become a standard design pattern in transformers:
the CLS token~\cite{dosovitskiy2021image,devlin2019bert} aggregates global semantics
for classification, memory tokens~\cite{burtsev2020memory} provide auxiliary storage
for long-range context, and task-specific query tokens in DETR~\cite{carion2020end}
attend to encoder features to produce object detections.
In vision,  Darcet \textit{et al.}~\cite{darcet2024registers} show that introducing register tokens helps ViTs absorb and discard outlier information, stabilizing the representations of the remaining patch tokens.
Building on this, Mamba-R~\cite{wang2024mambar} shows that Mamba-based vision models
also suffer from the same artifacts and benefit from the same register token fix~\cite{yang2025dinomamba}. Latest models such as Dino v2/v3~\cite{oquab2024dinov2,simeoni2025dinov3} and ViT-5~\cite{wang2026vit} have introduced registers as a default component.
RATS repurposes registers entirely: by placing them at the bottleneck of attention and training with DINO self-distillation, each register is forced to specialize to a distinct visual part, turning a cleanup mechanism into a part-level representation primitive.

\paragraph{Learned Latent Queries for Visual Token Aggregation.}
Set Transformer~\cite{lee2019set} and Perceiver/IO~\cite{jaegle2021perceiver,jaegle2022perceiver} established the $L{\to}N{\to}L$ cross-attention bottleneck pattern, but optimize supervised task objectives with no visual grouping semantics; Perceiver IO further entangles all latents through stacked $N{\times}N$ self-attention, preventing any latent from becoming an independent semantic unit.
Slot Attention~\cite{locatello2020object} instead uses learnable slot vectors 
that compete to represent scene elements via pixel reconstruction~\cite{burgess2019monet,greff2019multi,engelcke2020genesis};
DINOSAUR~\cite{seitzer2023dinosaur} and AdaSlot~\cite{fan2024adaslot} scale it to real images but remain at object level --- reconstruction yields object-level alpha masks, and hard slot competition tears apart semantically coherent regions.
Superpixel~\cite{mei2024spformer,jampani2018superpixel} groups tokens by low-level visual cues but strict spatial locality prevents capturing parts requiring global context; GroupViT~\cite{xu2022groupvit} achieves semantic grouping but requires image-text pairs.
In all cases, no prior work has demonstrated the ability to spontaneously 
surface semantic part-level representations through self-supervised learning alone.
RATS shows that the bottleneck paired with DINO self-distillation suffices 
to induce structured, transferable part-level primitives.

%% file: sec/3_method.tex
\section{Method}
\label{sec:method}
RATS makes a single architectural change to a Vision Transformer: every block's self-attention is augmented with a bottleneck attention that decomposes global aggregation into $N$ learnable register tokens, partitioned exclusively across the $H$ attention heads. We first review the preliminaries (\S\ref{subsec:prelim}), then describe the bottleneck block, its training, and the resulting register visualization (\S\ref{subsec:block}).

\subsection{Preliminaries}
\label{subsec:prelim}

\noindent\textbf{Vision Transformer.}
A Vision Transformer~\cite{dosovitskiy2021image} splits an input image into $L$ patches and adds a global classification \texttt{[CLS]} token, yielding a sequence $X\in\mathbb{R}^{(L+1)\times D}$ that passes through a stack of transformer blocks. With $H$ heads and per-head dimension $d=D/H$, the multi-head self-attention (MHSA)~\cite{vaswani2017attention} of a block is defined by
\begin{equation}
\label{eq:mhsa}
Q^{(h)}\!=\!\hat X W_Q^{(h)},\ \ K^{(h)}\!=\!\hat X W_K^{(h)},\ \ V^{(h)}\!=\!\hat X W_V^{(h)},\ \ \mathrm{Attn}^{(h)} \!=\! \mathrm{softmax}\!\left(\tfrac{Q^{(h)} K^{(h)\top}}{\sqrt d}\right)V^{(h)},
\end{equation}
where $\hat X=\mathrm{LN}(X)$ is layer normalization~\cite{ba2016layer}. The per-head outputs are concatenated and projected by a learned matrix $W_O\in\mathbb{R}^{D\times D}$, giving $\mathrm{MHSA}(\hat X)=\mathrm{Concat}(\mathrm{Attn}^{(1)},\ldots,\mathrm{Attn}^{(H)})\,W_O$. Both sub-layers employ pre-normalization and residual connections: $X\leftarrow X+\mathrm{MHSA}(\mathrm{LN}(X))$ and $X\leftarrow X+\mathrm{MLP}(\mathrm{LN}(X))$.

\noindent\textbf{DINO self-distillation.}
DINO~\cite{caron2021emerging} trains a student ViT $f_{\theta_s}$ to match the output of a teacher ViT $f_{\theta_t}$ on different augmented crops (views) of the same image: the teacher sees only the \emph{global} crops $\mathcal{V}^{g}$, while the student sees both \emph{global} and \emph{local} crops $\mathcal{V}=\mathcal{V}^{g}\cup\mathcal{V}^{l}$. Both networks share the same architecture --- a ViT backbone followed by an MLP projection head that maps the global \texttt{[CLS]} vector to a softmax distribution over $K$ learnable prototypes. The student is updated by gradient descent, while the teacher's weights are an exponential moving average (EMA) of the student's, $\theta_t \leftarrow \lambda\, \theta_t + (1-\lambda)\, \theta_s$, where $\lambda$ is the EMA decay rate. Writing $p_t(x)$ for the teacher's distribution on view $x$ after centering and sharpening at a low temperature $\tau_t$, and $p_s(x')$ for the student's distribution at a higher temperature $\tau_s$, the DINO objective is a cross-entropy summed over all teacher-global to student-view pairs:
\begin{equation}
\label{eq:dino}
\mathcal{L}_{\text{DINO}} \;=\; \sum_{x\,\in\, \mathcal{V}^{g}}\ \sum_{\substack{x'\,\in\, \mathcal{V} \\ x' \neq x}} -\, p_t(x)^{\!\top}\! \log p_s(x').
\end{equation}

\begin{figure}[!t]
\centering
\includegraphics[width=0.98\linewidth]{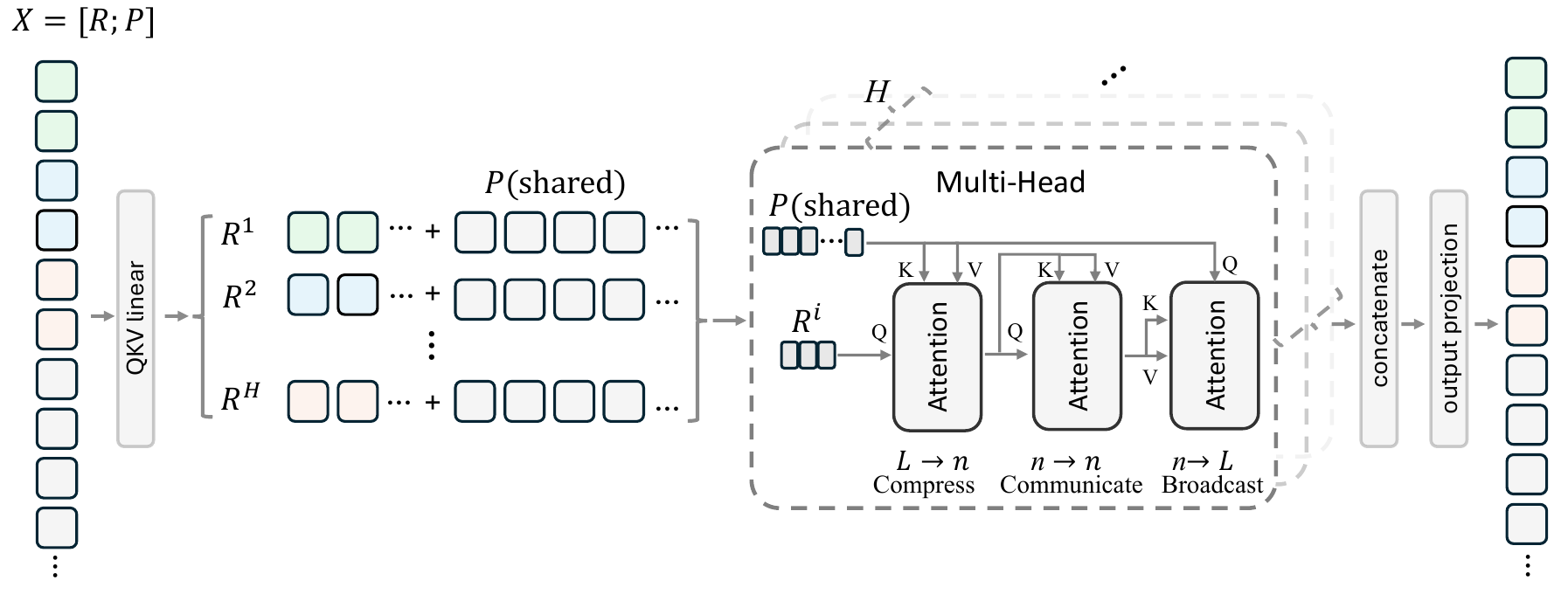}
\caption{\textbf{\textit{Register attention block}.} The shared QKV linear projects the input $X{=}[R;P]$ once. The register projections are partitioned into $H$ disjoint per-head slices $R^{1},\ldots,R^{H}$, while the patch projections $P$ are shared across all heads. Each head runs the three-step attention --- \emph{compress}, \emph{communicate}, \emph{broadcast} --- between its own register slice and the shared patches. Per-head outputs are concatenated and projected by the output projection $W_O$}
\label{fig:method}
\end{figure}

\subsection{RATS: Register Attention Transformers}
\label{subsec:block}
RATS introduces a bottleneck sub-block $\mathrm{R\text{-}Attn}(\cdot)$ that decomposes the \texttt{[CLS]} token into $N$ learnable \emph{register tokens}. Block input is $X =[R;\,P]\in\mathbb{R}^{(N+L)\times D}$, where $R\in\mathbb{R}^{N\times D}$ are register tokens and $P\in\mathbb{R}^{L\times D}$ are patch tokens. Rather than aggregating all patches into a single \texttt{[CLS]} token, $\mathrm{R\text{-}Attn}(\cdot)$ routes patch information through $N$ register tokens via an $L{\to}N{\to}N{\to}L$ bottleneck, each operating in a low-rank subspace (rank $\leq n = N/H$ per head). This \emph{compression} encourages each register to specialize to a proto-semantic, part-level region.

\begin{figure}[!h]
\centering
\includegraphics[width=0.98\linewidth]{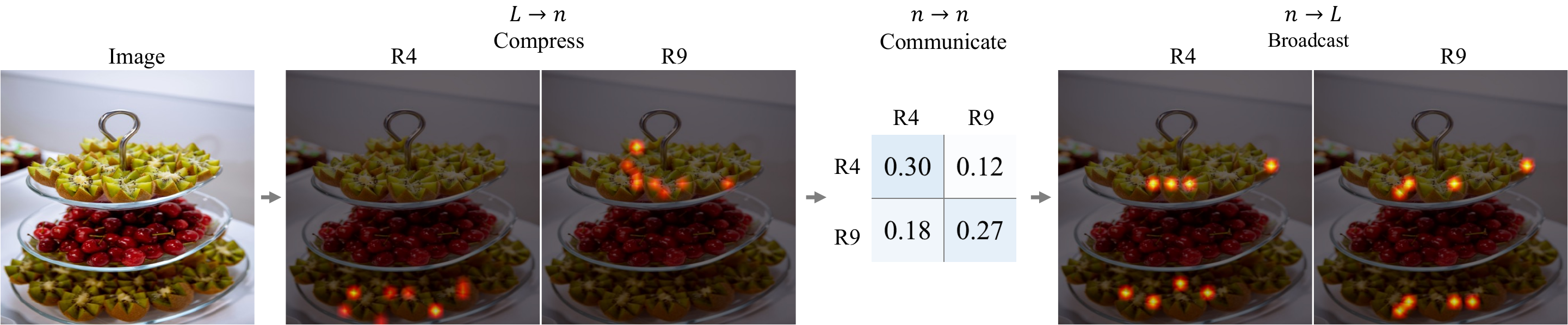}
\caption{\textbf{\textit{Visualization of the three-stage register attention block.}}
\emph{Compress}: registers $R_4$ and $R_9$ attend to distinct image regions (lower vs.\ upper tier). \emph{Communicate}: self-attention among registers lets co-occurring parts exchange context and bind into larger structures. \emph{Broadcast}: each register writes back to its own footprint, closing the $P{\to}R{\to}R{\to}P$ bottleneck and regrouping patches into proto-semantic regions.}

\label{fig:talk}
\end{figure}

\noindent\textbf{Single-head construction.}
Let $n$ denote the per-head register count and $d$ the per-head dimension
($n = N$ and $d = D$ when $H = 1$).
Applying projection matrices $W_Q,W_K,W_V\in\mathbb{R}^{D\times d}$
to the LayerNormed input $\hat X = \mathrm{LN}(X)$ yields
register projections $Q_R,K_R,V_R\in\mathbb{R}^{n\times d}$
and patch projections $Q_P,K_P,V_P\in\mathbb{R}^{L\times d}$.
The attention computation proceeds in three steps:
\begin{align}
\text{\emph{compress:}}\ \
  &\hat R \;=\; \mathrm{softmax}\!\left(\tfrac{Q_R K_P^{\!\top}}{\sqrt{d}}\right)V_P
  \;\in\;\mathbb{R}^{n\times d},
  \label{eq:s-compress}\\
\text{\emph{communicate:}}\ \
  &\bar R \;=\; \mathrm{softmax}\!\left(\tfrac{\hat R\,\hat R^{\!\top}}{\sqrt{d}}\right)\hat R
  \;\in\;\mathbb{R}^{n\times d},
  \label{eq:s-communicate}\\
\text{\emph{broadcast:}}\ \
  &O_P \;=\; \mathrm{softmax}\!\left(\tfrac{Q_P\,\bar R^{\!\top}}{\sqrt{d}}\right)\bar R
  \;\in\;\mathbb{R}^{L\times d}.
  \label{eq:s-broadcast}
\end{align}
\emph{Compress} aggregates patch information into registers, \emph{communicate} exchanges context among registers, and \emph{broadcast} writes the updated registers back to every patch. Figure~\ref{fig:talk} visualizes this three-stage pipeline. Intuitively, \emph{communicate} is self-attention among the registers: since each register already summarizes a coherent region, attending between them lets co-occurring parts exchange context and bind into larger structures---without it, patches would interact only indirectly through the compress--broadcast path, too weakly to capture patch--patch dependencies. Following \S\ref{subsec:prelim}, $\mathrm{R\text{-}Attn}$ replaces only the self-attention sub-layer; it keeps the same pre-norm residual connections and MLP, and the block outputs the updated tokens $[R;\,P]$.

\noindent\textbf{Multi-head extension.} In the Figure~\ref{fig:method},
with $H$ heads, per-head dimension $d = D/H$, and per-head register count
$n = N/H$, the $N$ registers are partitioned exclusively across heads.
Each head $h = 1,\ldots,H$ applies
Eqs.~\eqref{eq:s-compress}--\eqref{eq:s-broadcast} independently,
using its own register slice $R^{h} = R[hn:(h{+}1)n]$
and the $d$-dimensional projection slice of the shared $W_{QKV}$.
The outputs are assembled and projected:
\begin{equation}
\mathrm{R\text{-}Attn}(\hat X)
= \Bigl[\,O_R \;\Big|\; O_P\,\Bigr]\,W_O,
\label{eq:m-output}
\end{equation}
where $O_P = [O_P^{1},\ldots,O_P^{H}] \in \mathbb{R}^{L \times D}$ concatenates
heads along the feature axis, and $O_R \in \mathbb{R}^{N \times D}$ reassembles the
per-head register outputs $ R^{1},\ldots, R^{H}$ into the original
token and feature layout.

\noindent\textbf{Training.}
\label{subsec:train}
RATS is trained with the DINO objective (Eq.~\ref{eq:dino}) without modification. The input sequence $X^{(0)} = [R;\, \mathrm{PatchEmbed}(\mathbf{I}) + E_{\mathrm{pos}}]$ passes through $\mathcal{L}$ pre-norm blocks of $\mathrm{R\text{-}Attn}$ and MLP with residual connections, where positional encoding $E_{\mathrm{pos}}$ is applied only to the patch tokens. The mean of the final-layer register tokens $z = \tfrac{1}{N}\sum_{i=1}^{N} R_i^{}$ is fed to the DINO projection head $g_{\theta_s}$ as the global representation. The teacher shares the same architecture and is updated via EMA; no auxiliary region or part-level objective is used.

\paragraph{Register quality visualization.}
The similarity map between a register token and the patch tokens serves as a direct measure of the register's spatial selectivity: a highly localized and semantically coherent activation pattern indicates that the register has specialized to a visual part, while a diffuse or unstructured pattern suggests the register has not learned meaningful grouping. Accordingly, for each register $R_i$ we compute its cosine similarity with every patch token from the last block, yielding a score map $S_i \in \mathbb{R}^{h \times w}$. We rank all $N$ registers by their maximum activation and retain the top-$\tau$ most salient ones. The score maps are reshaped to the spatial grid and lightly smoothed with a Gaussian filter to reduce fragmentation, then each patch is assigned to its highest-scoring register via argmax to produce a segmentation map, as formalized in Eq.~\ref{eq:dominant_map} below. Figure~\ref{fig:teaser} (bottom) illustrates this pipeline and the resulting part-level similarity maps.
\begin{equation}
\label{eq:dominant_map}
Seg(u,v) \;=\; \arg\max_{i\,\in\,\mathcal{T}}\;
             \bigl(G_\sigma \ast S_i\bigr)(u,v),
\qquad
S_i(u,v) \;=\; \frac{\langle R_i,\,P_{u,v}\rangle}{\|R_i\|\,\|P_{u,v}\|}.
\end{equation}
where $\mathcal{T}$ indexes the top-$\tau$ registers ranked by $\max_{u,v} S_i(u,v)$, and $G_\sigma$ is a $2$-D isotropic Gaussian kernel with standard deviation $\sigma$.

\begin{figure}[!h]
\centering
\includegraphics[width=0.98\linewidth]{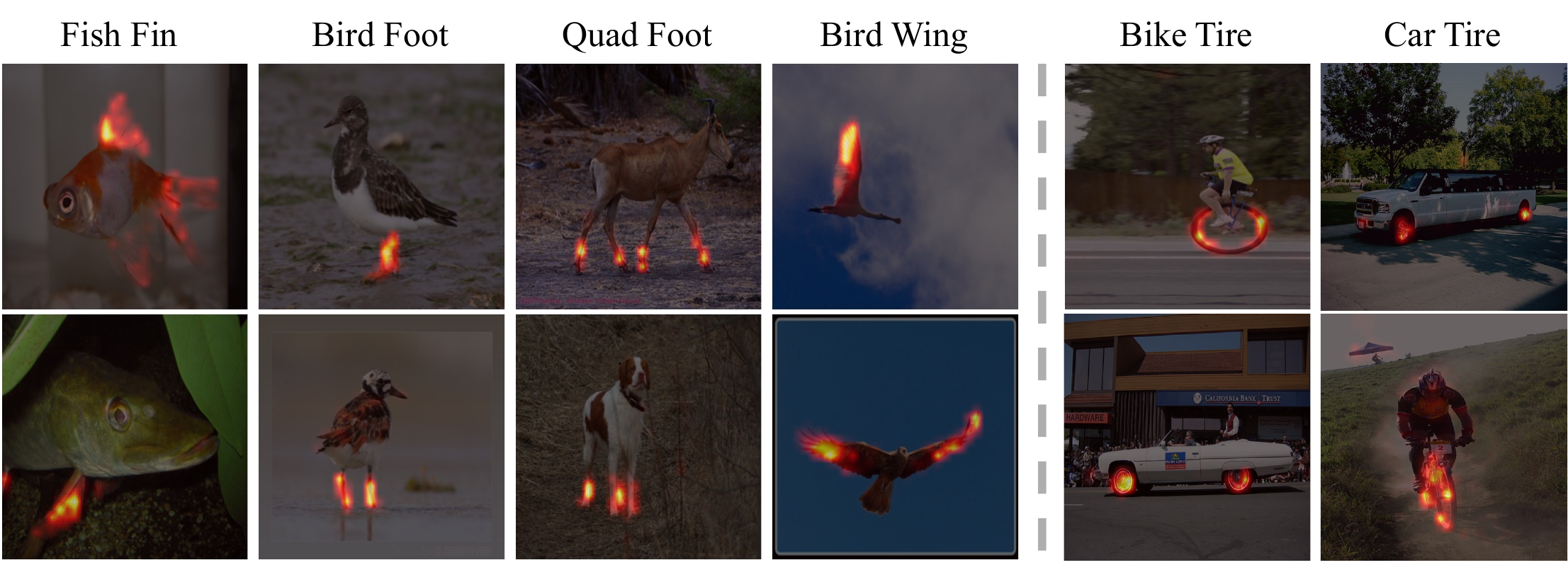}
\caption{\textbf{\textit{Qualitative visualisation of individual dictionary entries.}}  Each column shows attention overlays of a dictionary entry on two test images. \emph{Left}: a single entry fires on the same part across different instances within a super-category. \emph{Right}: Semantically similar dictionary entries (e.g., wheel-like entries) generalize across different super-categories.}
\label{fig:dict_qualitative}

\end{figure}

\paragraph{From Registers to a Part Dictionary}
\label{subsec:dict_concept}
Each register attends to a specific region rather than summarizing the whole image, so its activation is a localized \emph{part feature}. Because parts recur across images, these features cluster into a compact \emph{dictionary} of part entries shared across the dataset. A standard ViT cannot expose such a dictionary: its single global \texttt{[CLS]} token provides no per-region tokens to aggregate. We construct it and evaluate its consistency and cross-category transfer quantitatively in \S\ref{subsec:dict}; Figure~\ref{fig:dict_qualitative} shows example entries.

%% file: sec/4_exp.tex
\section{Experiments}

\subsection{Implementation details.}

We pretrain two model scales, RATS-S/B, on ImageNet-1k~\cite{deng2009imagenet} at $224{\times}224$ using the DINO recipe without modification. We evaluate on five benchmarks: COCO 2017 val~\cite{lin2014microsoft}, ADE20K~\cite{zhou2017scene}, ImageNet val, ImageNet-S$_{919}$~\cite{gao2022luss}, and PartImageNet~\cite{he2022partimagenet}. Since ImageNet lacks segmentation masks and COCO instance annotations cover only a small fraction of each image, we adopt SAM~2.1~\cite{kirillov2023segment,ravi2024sam2} generated masks as proxy ground truth for these two datasets, and use native annotations for the remaining three. We report many-to-one mIoU (M2O) and ARI as primary metrics. For downstream transfer, we fine-tune with Mask2Former on ADE20K semantic segmentation and COCO detection/instance segmentation, comparing against DINO ViT-S/16 under identical recipes. Full details are provided in Appendix~\ref{app:impl}.

\subsection{Region Part Segmentation}
\begin{figure}[!h]
\vspace{-0.5cm}
\centering
\includegraphics[width=1.0\linewidth]{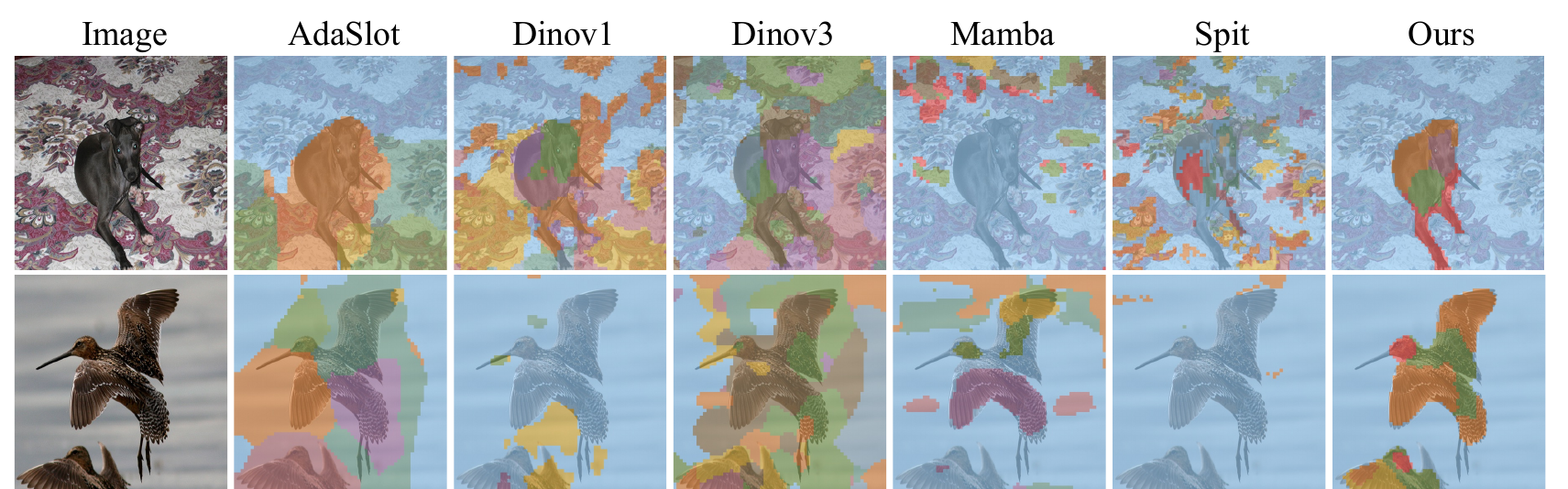}
\caption{\textbf{\textit{Qualitative comparison of part segmentation.}} RATS yields cleaner foreground--background separation and more part-coherent regions than the baselines. Extended comparison in Appendix~\ref{app:qualitative}.}
\label{fig:qualitative_compare}
\vspace{-0.4cm}
\end{figure}

We compare RATS against representative self-supervised ViT backbones (DINOv1~\cite{caron2021emerging}, DINOv3~\cite{simeoni2025dinov3}), Mamba~\cite{gu2023mamba,zhu2024vision,Adventurer} Backbone pretrained with register~\cite{yang2025dinomamba}, superpixel-/region-oriented methods (SPFormer~\cite{mei2024spformer}, SPiT~\cite{aasan2024spit}), and a recent slot-based model (AdaSlot~\cite{fan2024adaslot}). We report many-to-one mIoU (\%) on five benchmarks and the average ARI. As shown in Table~\ref{tab:main_seg}, RATS establishes a new state of the art on every dataset and improves the average mIoU by \textbf{+12} over the strongest baseline (AdaSlot), consistent with the design goal of the register bottleneck to produce semantically coherent, part-aware regions. Figure~\ref{fig:qualitative_compare} corroborates this qualitatively: where the baselines either fragment the foreground or leak into the background, RATS produces clean, grounded regions.

\begin{table}[h]
\centering
\tablestyle{3pt}{1.1}
{%
\begin{tabular}{l|c|c|ccccc|c|c}
Method & Backbone & \#Register & COCO & ADE20K & IN & IN-S$_{919}$ & PartIN & Avg & ARI \\
\shline
\multicolumn{10}{l}{\textit{\textcolor{gray}{Standard ViT}}} \\
DINOv1  & ViT-S/16 & 6  &  8.53 & 15.43 & 11.21 & 48.28 & 25.58 & 21.81 & 8.7 \\
DINOv3  & ViT-S/16 & 6  &  8.17 & 13.69 & 11.15 & 50.25 & 27.67 & 22.19 & 11.1 \\
DINOv1  & ViT-B/16 & 12 &  8.55 & 15.87 & 11.05 & 46.94 & 25.82 & 21.65 & 7.9 \\
DINOv3  & ViT-B/16 & 12 & 10.49 & 18.46 & 14.13 & 56.43 & 32.57 & 26.41 & 12.3 \\
\hline
\multicolumn{10}{l}{\textit{\textcolor{gray}{Mamba with Register}}} \\
DINOv1  & Mam.R-B/16 & 12  &  10.07 & 18.95 & 13.44 & 55.54 & 30.11 & 25.62 & 12.8 \\
\hline
\multicolumn{10}{l}{\textit{\textcolor{gray}{Superpixel}}} \\
SPFormer  & ViT-S/16 & 6  &  8.37 & 14.65 & 12.10 & 51.92 & 27.21 & 22.85 & 11.5 \\
SPiT       & ViT-B/16 & 12 &  8.41 & 12.89 & 12.37 & 53.47 & 27.83 & 23.00 & 8.5 \\
\hline
\multicolumn{10}{l}{\textit{\textcolor{gray}{Slot}}} \\
AdaSlot           & ViT-B/16 & 33 & 12.31 & 23.87 & 15.49 & 54.83 & 30.74 & 27.45 & 19.3 \\
\hline
\multicolumn{10}{l}{\textit{\textcolor{gray}{Ours}}} \\
\rowcolor{cvprblue!5}
DINOv1             &RATS-S/16 & 6   & 9.91 & 17.02 & 13.97 & 58.84 & 33.93 & 26.73 & 20.4 \\
\rowcolor{cvprblue!5}
DINOv1             & RATS-B/16 & 12  & 12.99 & 21.45 & 17.31 & 67.89 & 41.06 & 32.14 & 22.0 \\

\rowcolor{cvprblue!5}
DINOv1             & RATS-B/16 & 33  & 16.89 & 27.52 & 21.32 & 73.62 & 46.94 & 37.26 & 21.1 \\
\rowcolor{cvprblue!15}
DINOv1      & RATS-S/16 & 96  & \textbf{17.31} & \textbf{26.98} & \textbf{21.87} & \textbf{73.18} & \textbf{46.43} & \textbf{37.15} & \textbf{21.7} \\
\rowcolor{cvprblue!15}
DINOv1     & RATS-B/16 & 192 & \textbf{19.32} & \textbf{30.77} & \textbf{23.47} & \textbf{75.39} & \textbf{48.77} & \textbf{39.54} & \textbf{19.6} \\

\end{tabular}%
}
\caption{\textbf{\textit{part segmentation.}} M2O mIoU (\%) on five benchmarks and mean ARI. RATS outperforms self-supervised, superpixel, and slot-based baselines at matched token budgets, and surpasses AdaSlot with fewer tokens.}
\label{tab:main_seg}

\end{table}

\subsection{Downstream Transfer}

The preceding experiments demonstrate that RATS's register bottleneck discovers semantically coherent part-level representations. A natural question is whether the register tokens can serve as input queries for query-based decoders such as Mask2Former~\cite{cheng2022masked}. To test this, we add the $L{\to}N{\to}N{\to}L$ bottleneck to a standard ViT and use the resulting register tokens as queries for Mask2Former, fine-tuning on ADE20K semantic segmentation and COCO 2017 detection/instance segmentation. We compare against the same architecture initialized from a DINO ViT-S/16 checkpoint under identical recipes (full details in Appendix~\ref{app:impl}).
As shown in Table~\ref{tab:downstream_transfer}, RATS pretraining consistently outperforms the DINO baseline on both ADE20K and COCO. The parenthesized values report results when replacing only the backbone, while keeping the original independently learned queries unchanged, thereby isolating the backbone contribution. This reveals that RATS's gains stem from two complementary sources: the bottleneck training produces stronger backbone features, while the register tokens provide part-level semantic representations that serve as superior query inputs.

\begin{table}[h]
\centering
\tablestyle{5pt}{1.1}
{%
\begin{tabular}{l|c|ccc|cc}
 & & \multicolumn{3}{c|}{ADE20K (sem.\ seg.)} & \multicolumn{2}{c}{COCO 2017 (det.\ \& inst.\ seg.)} \\
Pretraining & \#queries & aAcc (\%) & \textbf{mIoU (\%)} & mAcc (\%) & AP$^{\text{b}}$ (\%) & \textbf{AP$^{\text{m}}$ (\%)} \\
\shline
DINO                       & 100   & 81.37 & 45.41 & 58.23 & 41.1 & 37.9 \\
\rowcolor{cvprblue!15}
\textbf{RATS}               &  96 (100 init queries) & \textbf{81.61} {\scriptsize(81.47)} & \textbf{46.52} {\scriptsize(45.83)} & \textbf{59.37} {\scriptsize(60.44)} & \textbf{41.2} & \textbf{38.1} \\
\end{tabular}%
}
\caption{\textbf{\textit{Downstream transfer to ADE20K and COCO 2017.}} Fine-tuning Mask2Former~\cite{cheng2022masked} with ViT-S/16 backbone, initialized from DINO or RATS pretraining (both 100 epochs on ImageNet-1k). Parenthesized values denote results when replacing only the backbone, while keeping the standard learned queries unchanged.}
\label{tab:downstream_transfer}
\vspace{-0.5cm}
\end{table}

\subsection{Dictionary Analysis}
\label{subsec:dict}

RATS's registers produce strong part-level features; we ask whether these can be aggregated into a dictionary of consistent, shared parts per category, and evaluate this on PartImageNet.

\begin{figure}[h]
\centering
\begin{minipage}[c]{0.53\linewidth}
\centering
\tablestyle{3pt}{1.1}
{\small
\begin{tabular}{l|cc|c|c}
Method & mAP $\uparrow$ & R@1 $\uparrow$ & valPw $\uparrow$ & util\% $\uparrow$ \\
\shline
Random baseline        & 0.16 & 0.16 & 0.150 & -- \\
Post-hoc K-means       & \textbf{0.56} & \textbf{0.93} & 0.306 & \textbf{100} \\
\rowcolor{cvprblue!15}
\textbf{Learned codebook}       & 0.48 & 0.83 & 0.317 & 62 \\
Learned codebook (L)   & 0.45 & 0.82 & \textbf{0.355} & 10 \\
\end{tabular}%
}
\end{minipage}\hfill
\begin{minipage}[c]{0.42\linewidth}
\centering
\includegraphics[width=\linewidth]{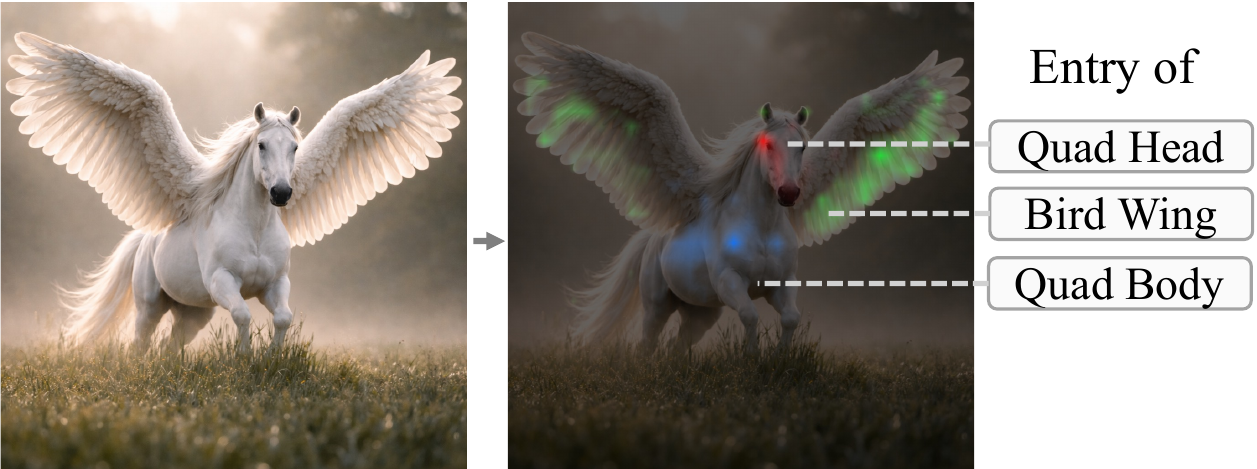}
\end{minipage}
\caption{\textbf{\textit{Codebook construction and compositional generalisation.}}
\emph{Left}: retrieval and purity metrics under different codebook constructions.
\emph{Right}: a \textit{Pegasus} (unseen composite of horse and bird wings) is decomposed by three independently learned 
entries, supporting zero-shot decomposition of novel compositions.}
\label{fig:codebook_ood}

\end{figure}

\noindent\textbf{Codebook construction.}
A register that activates on a spatial region attends to that region in its attention map, yielding a localised part feature. We compare four codebook constructions in Fig.~\ref{fig:codebook_ood} (left). 
\emph{Post-hoc K-means} serves as an upper bound: it extracts register features within ground-truth part masks and clusters them offline. \emph{Learned codebook} randomly initialises a set of entries and trains them on ImageNet by minimising the distance between each register and its nearest entry; ground-truth masks are used only afterwards to label each entry with its dominant part. The learned codebook trails K-means on retrieval by a small margin while delivering higher per-entry purity, confirming that RATS's register features 
are already well-structured for part discovery. Enlarging the vocabulary (\emph{Learned codebook (L)}) further raises purity but wastes $90\%$ of entries, exposing a purity--utilisation trade-off.
We assess the dictionary along two axes: image-level retrieval (mAP, R@1) under a bag-of-entries representation, and per-entry weighted purity with utilisation (valPw, util\%); see Appendix~\ref{app:dict_eval} for definitions.

\noindent\textbf{Qualitative verification.}
Figure~\ref{fig:dict_qualitative} shows that individual entries fire consistently on the same part within a super-category (left) and that functional concepts such as ``wheel'' transfer across super-categories (right). Fig.~\ref{fig:codebook_ood} (right) further shows that a \textit{Pegasus}---an unseen composite of horse and bird wings---is decomposed by independently learned entries, supporting zero-shot compositional generalisation. Additional analysis is in Appendix~\ref{app:dict_eval}.

\subsection{Ablation Studies}

\noindent\textbf{Design Choices for Register-Based Token Aggregation.}
We ablate three axes: (i) per-head partition vs shared registers (all heads attend to the full register set); (ii) parameter-free vs learned communicate projections; (iii) RATS vs a Perceiver-IO style structure~\cite{jaegle2022perceiver} that compresses patches into 80 latents, processes them with 10 self-attention layers, then decodes via 96 output queries. Table~\ref{tab:design_space} reports cost (NVIDIA A5000 GPU, 512$\times$512, batch=16) and quality averaged over five datasets at top-$\tau{=}6$. RATS achieves lower latency and memory than ViT-S/16 at identical parameter count with improved segmentation quality, Pareto-dominating all design alternatives.
\begin{table}[!h]
\centering
\tablestyle{3pt}{1.1}
{%
\begin{tabular}{l|cccc|ccc}
Model & Params & Mem & Latency & rel.~cost & M2O $\uparrow$ & O2O $\uparrow$ & ARI $\uparrow$ \\
\shline
ViT-S/16 (DINO) & 21.67~M & 1.08~GB & 172.8~ms & 2.12$\times$ & 21.8 & 19.3 & 8.7 \\
\rowcolor{cvprblue!15}
RATS (Baseline) & 21.70~M & 0.49~GB & 81.7~ms & 1.00$\times$ & \textbf{26.6} & \textbf{23.1} & \textbf{19.2} \\
RATS + learned comm & 22.59~M & 0.49~GB & 82.6~ms & 1.01$\times$ & 25.5 & 21.9 & 17.1 \\
RATS, shared & 21.70~M & 0.52~GB & 91.7~ms & 1.12$\times$ & 25.0 & 21.2 & 15.1 \\
Perceiver-IO style & 21.73~M & 0.31~GB & 12.4~ms & 0.15$\times$ & 20.2 & 18.6 & 10.0 \\
\end{tabular}%
}
\caption{\textbf{\textit{Ablation on attention design choices.}} 
Each row varies one design axis from RATS (Baseline). Perceiver-IO minimises 
cost but sacrifices spatial detail.}
\label{tab:design_space}
\vspace{-0.5cm}
\end{table}

\noindent\textbf{Number of per-head registers $n$ and over-segmentation.}
The register bottleneck compresses patch information into $N{=}n{\times}H$ region tokens, directly controlling grouping granularity. We ablate $n \in \{8, 16, 24, 32\}$ on RATS-S ($H{=}6$) at top-$\tau{=}40$ and report both many-to-one (M2O) and one-to-one (O2O) mIoU. M2O reflects clustering quality, while a large M2O--O2O gap indicates that one semantic class is fragmented across multiple registers (over-segmentation). As shown in Table~\ref{tab:ablation_N}, $n{=}16$ achieves the best M2O, while smaller $n$ only leads to a slight drop. Larger $n$ tends to increase over-segmentation, whereas smaller $n$ produces coarser but more complete regions.

\begin{table}[h]
\centering
\tablestyle{2.5pt}{1.1}
{%
\begin{tabular}{l|cccccc|cccccc}
 & \multicolumn{6}{c|}{mIoU M2O (\%) $\uparrow$} & \multicolumn{6}{c}{mIoU O2O (\%) $\uparrow$} \\
$n$ & COCO & ADE20K & ImageNet & IN-S919 & PartIN & Avg & COCO & ADE20K & ImageNet & IN-S919 & PartIN & Avg \\
\shline
8   & 15.67 & 23.97 & 19.96 & 71.47 & 45.71 & 35.36 & 14.70 & 21.03 & 17.45 & \textbf{43.10} & \textbf{32.89} & \textbf{25.83} \\
\rowcolor{cvprblue!15}
\textbf{16}  & 16.77 & \textbf{26.59} & \textbf{21.37} & \textbf{72.72} & 45.67 & \textbf{36.62} & \textbf{15.81} & 22.76 & \textbf{18.34} & 40.28 & 30.82 & 25.60 \\
24  & \textbf{16.89} & 26.46 & 21.19 & 72.30 & 46.14 & 36.60 & 15.70 & \textbf{22.85} & \textbf{18.34} & 39.65 & 30.87 & 25.48 \\
32  & 16.13 & 25.19 & 20.82 & 71.81 & \textbf{46.22} & 36.03 & 15.65 & 22.25 & 18.23 & 39.79 & 31.53 & 25.49 \\
\end{tabular}%
}
\caption{\textbf{\textit{Effect of per-head register count $n$ and over-segmentation.}} M2O vs.\ O2O mIoU (\%) on RATS-S. Higher M2O reflects better clustering; a large M2O--O2O gap indicates over-segmentation. $n{=}16$ achieves the best M2O, while $n{=}8$ trades absolute M2O for higher O2O on part-centric classes.}
\label{tab:ablation_N}

\end{table}

\noindent\textbf{Top-$\tau$ region ablation.}
At inference, we rank registers by activation magnitude and select the top-$\tau$ to form the segmentation map. Table~\ref{tab:ablation_topn} reveals a coverage-purity trade-off: increasing $\tau$ from 10 to 80 improves mIoU by up to 9.2\% on IN-S919, but ARI consistently decreases (e.g., $-$2.9\% on IN-S919), since lower-ranked registers attend to less coherent areas. The metrics saturate beyond $\tau{\approx}50$, reflecting a natural ranking among visual primitives learned by the bottleneck.
\begin{table}[!h]
\centering
\tablestyle{3pt}{1.1}
{%
\begin{tabular}{l|ccccc|ccccc}
 & \multicolumn{5}{c|}{mIoU (M2O, \%) $\uparrow$} & \multicolumn{5}{c}{ARI $\uparrow$} \\
Top\_$\tau$ & COCO & ADE20K & ImageNet & IN-S919 & PartIN & COCO & ADE20K & ImageNet & IN-S919 & PartIN \\
\shline
10 & 11.76 & 20.07 & 15.99 & 64.00 & 37.72 & 20.30 & 23.55 & 23.22 & \textbf{19.74} & \textbf{24.00} \\
\rowcolor{cvprblue!5}
30 & 15.99 & 25.58 & 20.66 & 71.88 & 44.67 & \textbf{23.01} & 24.77 & 24.74 & 17.78 & 21.22 \\
50 & 17.07 & 26.87 & 21.70 & 73.01 & 46.18 & 22.88 & 24.44 & 24.39 & 17.05 & 20.21 \\
\rowcolor{cvprblue!15}
80 & \textbf{17.30} & \textbf{26.98} & \textbf{21.87} & \textbf{73.18} & \textbf{46.43} & 22.84 & 24.39 & 24.29 & 16.89 & 20.11 \\
\end{tabular}%
}
\caption{\textbf{\textit{Effect of top-$\tau$ selection.}} Larger $\tau$ improves spatial coverage (higher mIoU) but dilutes clustering purity (lower ARI), revealing a coverage-purity trade-off. }
\label{tab:ablation_topn}

\end{table}

\begin{table}[h]
\centering
\tablestyle{3pt}{1.1}
{%
\begin{tabular}{l|ccccc}
 & \multicolumn{5}{c}{mIoU (M2O, \%) $\uparrow$} \\
Loss Scope & COCO & ADE20K & ImageNet & IN-S919 & PartIN \\
\shline
Per-head & 15.38 & 21.82 & 18.47 & 67.20 & 39.76 \\
\rowcolor{cvprblue!15}
\textbf{All-head} & \textbf{16.77} & \textbf{26.59} & \textbf{21.37} & \textbf{72.72} & \textbf{45.67} \\
\end{tabular}%
}
\caption{\textbf{\textit{DINO loss placement.}} mIoU (M2O). \textit{Per-head}: each attention head's register tokens receive a separate DINO distillation loss. \textit{All-head}: a single shared DINO loss over all registers.}
\label{tab:ablation_loss}
\vspace{-0.5cm}
\end{table}

\noindent\textbf{DINO loss placement.}
We investigate whether each attention head benefits from an independent DINO loss to learn distinct patterns, or whether a single loss over all registers suffices. As shown in Table~\ref{tab:ablation_loss}, 
applying a separate DINO loss per head degrades mIoU, indicating that a shared loss over all registers is more effective.

\begin{figure}[!h]
\centering
\includegraphics[width=0.98\linewidth]{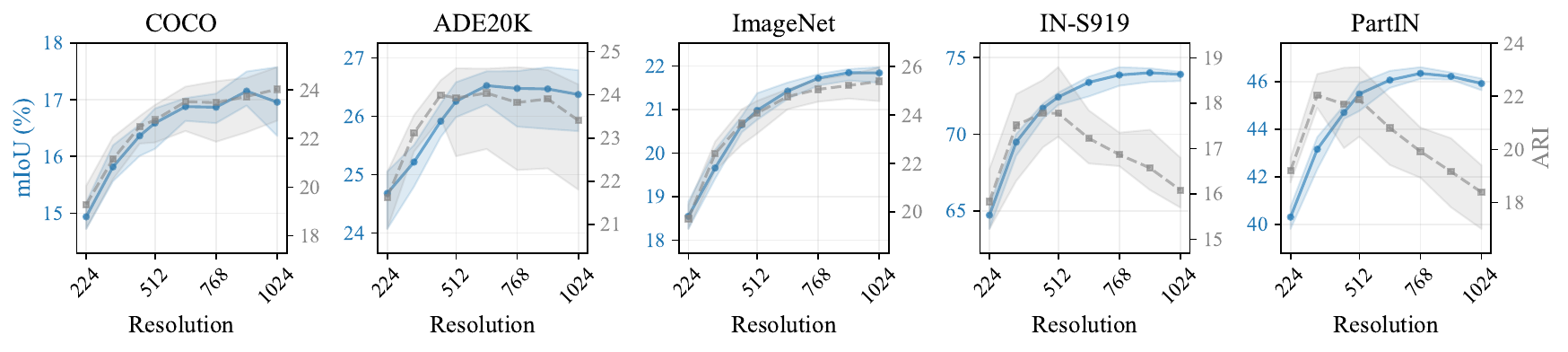}
\caption{\textbf{\textit{Effect of inference resolution (RATS-S).}} mIoU (M2O, \%, blue, left axis) and ARI (gray, right axis) across resolutions 224--1024. Lines show the mean over epoch 140--200; shaded bands show the min--max range. mIoU improves up to 640 then saturates; ARI peaks at intermediate resolutions, revealing a resolution--coherence trade-off.}
\label{fig:resolution}

\end{figure}

\noindent\textbf{Effect of inference resolution.}
DINO-based models naturally support variable-resolution inputs at inference. Using RATS-S as a representative model, we sweep resolution from 224 to 1024 and report the mean across checkpoints at epoch 140--200, with the shaded band showing the min--max range (Figure~\ref{fig:resolution}). mIoU increases steadily up to 640, as finer patch grids provide more spatial detail for the register bottleneck. Beyond 640, gains saturate on most datasets while ARI begins to decline on ADE20K and PartIN, indicating that excessive patch counts dilute attention coherence.

\begin{figure}[!h]
\centering
\includegraphics[width=0.95\linewidth]{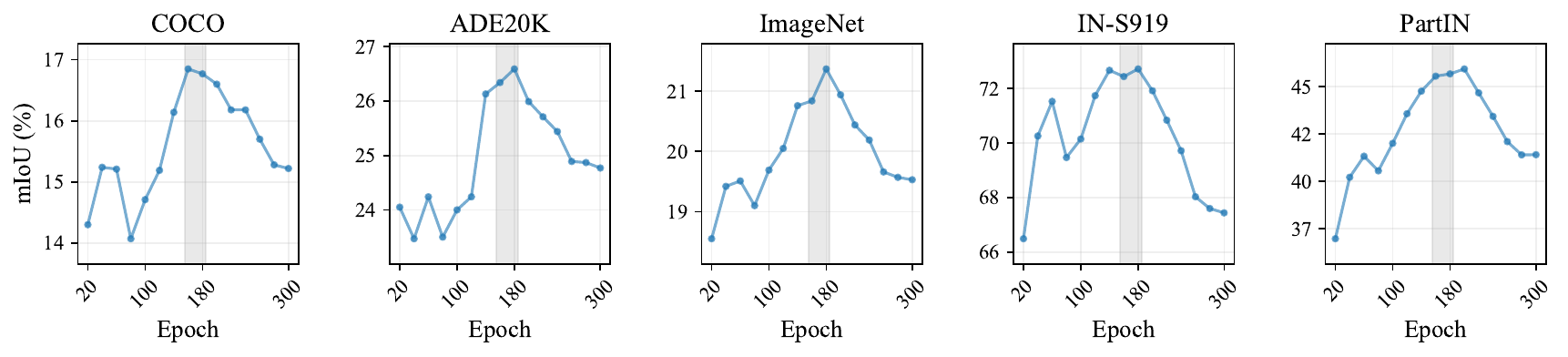}
\caption{\textbf{\textit{Training dynamics (RATS-S).}} mIoU (M2O, \%) across epochs 20--300 on five benchmarks. The gray band marks the optimal window (epoch 160--180). Performance degrades with continued training beyond this point, indicating diminishing returns from over-distillation. Peak values are annotated for each dataset.}
\label{fig:training_dynamics}

\end{figure}

\noindent\textbf{Training dynamics.}
We track mIoU of RATS-S across training epochs to understand when region structure emerges (Figure~\ref{fig:training_dynamics}). Performance rises rapidly in early training, reaching $\sim$85\% of peak by epoch 20. All five datasets peak around epoch 160--180, then gradually decline through epoch 300. This degradation indicates diminishing register diversity. The gray band marks the optimal training window of 160--180 epochs.

%% file: sec/5_con.tex
\section{Conclusion}

We present RATS, a Vision Transformer architecture built on a register-token bottleneck that enables self-supervised models to spontaneously discover part-level representations without any part annotation or auxiliary loss. Experiments show that this simple architectural design surpasses existing methods on part segmentation and provides stronger features and query initialization for downstream dense prediction. The visual dictionary extracted from the trained registers further reveals structured semantic organization, including within-super-category part consistency and cross-category taxonomic proximity. Our results demonstrate that architectural inductive bias alone can serve as an effective prior for part discovery, providing a promising foundation for fine-grained recognition, interpretable visual analysis, and broader compositional visual understanding.

%% file: sec/6_appendix.tex
\section*{Appendix}
\addcontentsline{toc}{section}{Appendix}

\section{Limitations}
\label{app:limitations}

RATS has two main limitations. \textbf{(i) Low-rank attention.} To allow registers to spontaneously acquire part-segmentation ability, we adopt the $L{\to}N{\to}N{\to}L$ structure. Because the $L{\to}N$ stage compresses information through only $n{=}N/H$ registers per head, the resulting attention has rank at most $n$, which is lower than the rank of standard $L{\to}L$ self-attention. For tasks that require maintaining a high-rank representation, such as semantic segmentation and object detection, we therefore introduce an additional parallel $L{\to}L$ branch during fine-tuning to recover full-rank capacity (Appendix.~\ref{app:impl}). \textbf{(ii) Mismatch with human-defined semantics.} The grouping discovered by RATS is obtained without any annotation, so its part boundaries are determined by visual co-occurrence rather than by human-defined taxonomies. Compared with supervised segmentation, which targets human-labelled categories, the resulting regions may not always align with the parts a human would name, and the segmentation should be interpreted as an emergent visual decomposition rather than a label-faithful one.

\section{Extended qualitative comparison}
\label{app:qualitative}

\begin{figure}[!h]
\centering
\includegraphics[width=\linewidth]{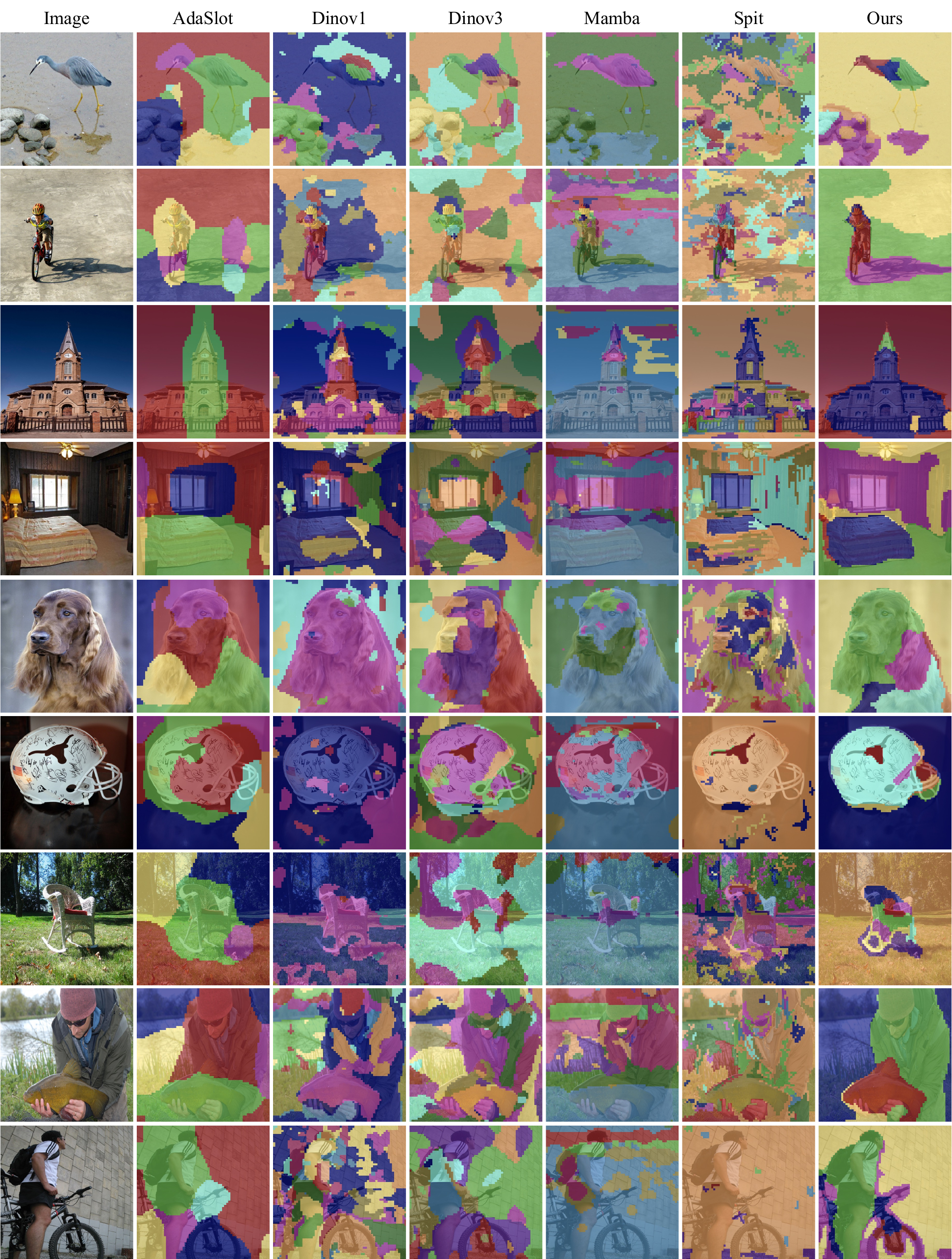}
\caption{\textbf{\textit{Extended qualitative comparison on part region segmentation.}} Same column layout as Figure~\ref{fig:qualitative_compare} in the main paper. RATS produces consistent foreground--background separation and part-coherent regions across diverse natural images.}
\label{fig:qualitative_compare_app}
\end{figure}

Figure~\ref{fig:qualitative_compare_app} extends the qualitative comparison of Figure~\ref{fig:qualitative_compare} (main paper) with nine additional images spanning animals (bird, cocker spaniel), vehicles (bicycle, mountain bike), indoor/outdoor scenes (cathedral, bedroom), and challenging compositions (text-covered helmet, sculpture, person with prop). The same column layout is used: input image followed by AdaSlot~\cite{fan2024adaslot}, DINOv1~\cite{caron2021emerging}, DINOv3~\cite{simeoni2025dinov3}, Mamba, SPiT~\cite{aasan2024spit}, and RATS. Across all nine images, RATS consistently delivers cleaner foreground--background separation and more part-coherent regions than the baselines.

\section{Dictionary Evaluation and Structure}
\label{app:dict_eval}

This appendix expands on the dictionary analysis reported in Section~\ref{subsec:dict} of the main paper. We first describe the image representation, metrics, and codebook variants used in Figure~\ref{fig:codebook_ood}, then present the structural analysis behind the qualitative claims in Figure~\ref{fig:dict_qualitative}.

\begin{figure}[h]
\centering
\includegraphics[width=0.8\linewidth]{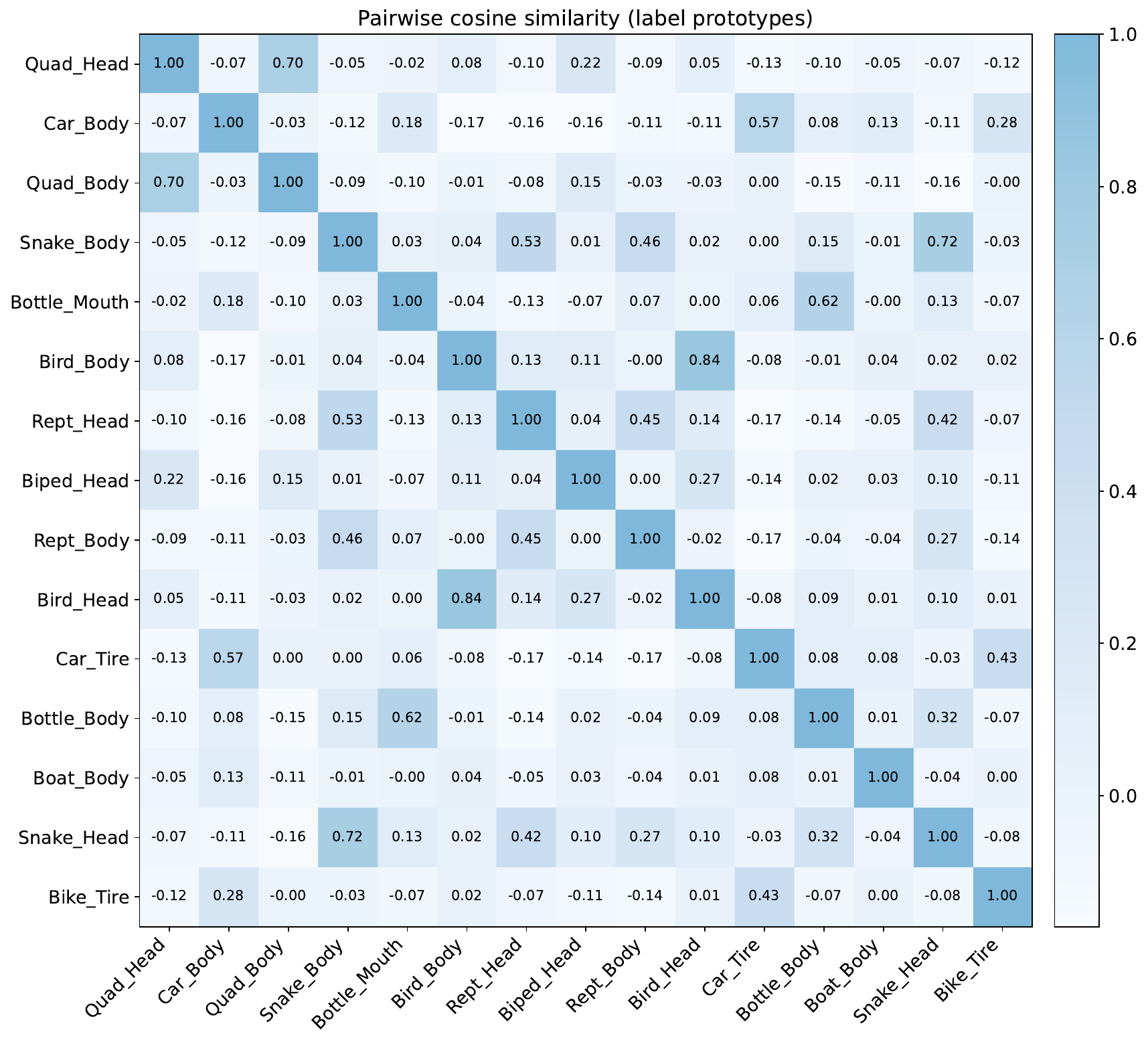}
\caption{\textbf{\textit{Pairwise cosine similarity between label prototypes}.} Each row/column is the averaged dictionary entry of one dominant part label. Three structures emerge---within-object coherence, taxonomic proximity, and functional analogy---analysed in Sec.~\ref{app:dict_eval}.}
\label{fig:label_distance}
\end{figure}

\subsection{Qualitative analysis of the dictionary}

\begin{figure}[h]
\centering
\includegraphics[width=\linewidth]{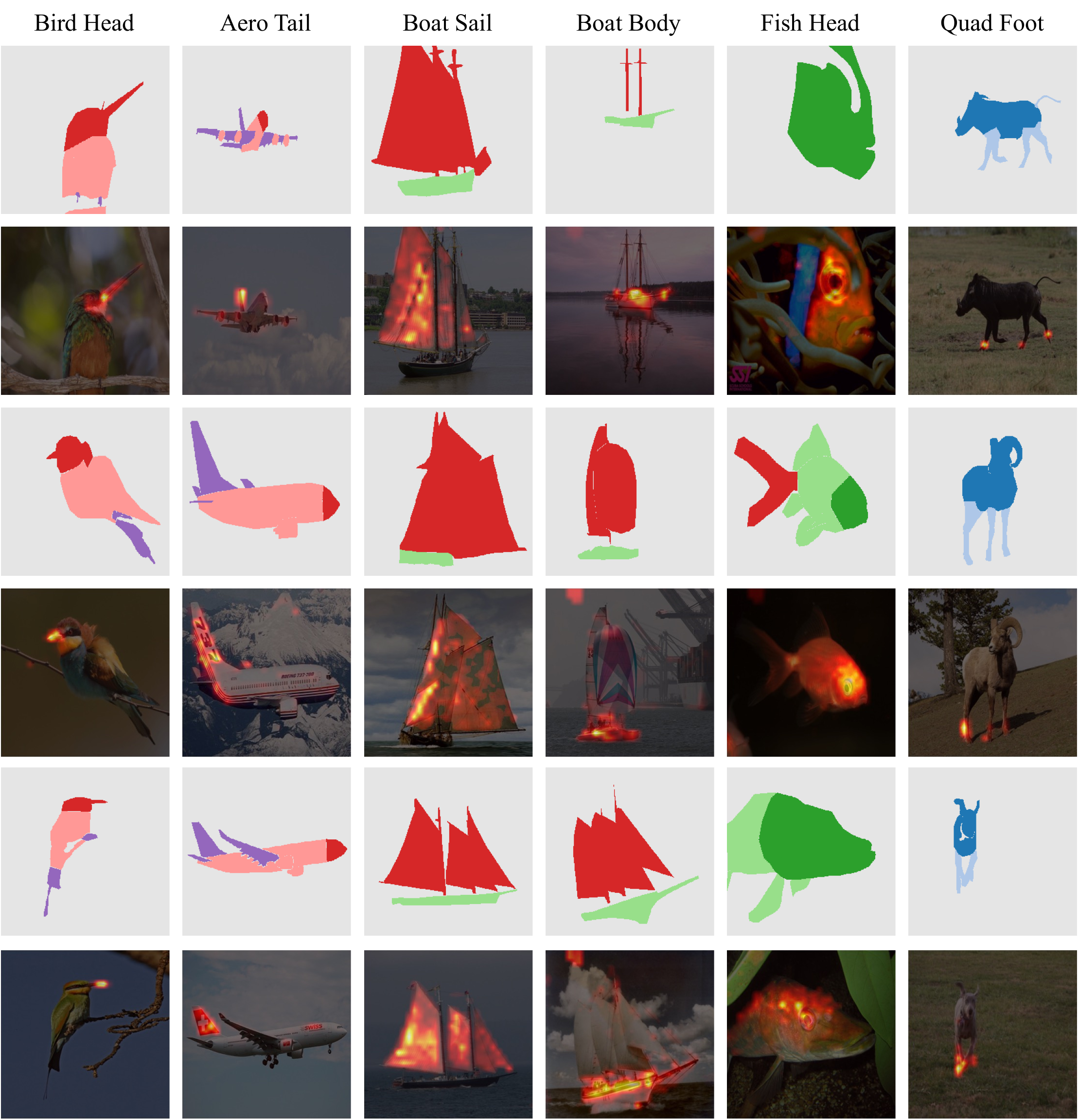}
\caption{\textbf{\textit{Within-super-category consistency of individual dictionary entries.}} Each column corresponds to one dictionary entry; each of the three (ground-truth part segmentation, register-attention overlay) pairs shows a different PartImageNet class from the same super-category. The attention consistently aligns with the annotated part boundary across instances, and the six entries together span both biological (\textit{Bird\_Head}, \textit{Fish\_Head}, \textit{Quad\_Foot}) and rigid man-made (\textit{Aero\_Tail}, \textit{Boat\_Sail}, \textit{Boat\_Body}) super-categories, despite no part-level supervision being used during training.}
\label{fig:dict_qualitative_app}
\end{figure}

Figure~\ref{fig:dict_qualitative_app} shows register-attention overlays alongside ground-truth segmentations for six dictionary entries, each evaluated on three images from different classes within the same super-category. Each entry consistently highlights the target part across varied instances, closely matching human-annotated boundaries, and the illustrated entries span both deformable biological parts (\textit{Bird\_Head}, \textit{Fish\_Head}, \textit{Quad\_Foot}) and rigid man-made parts (\textit{Aero\_Tail}, \textit{Boat\_Sail}, \textit{Boat\_Body}), confirming that a single learned vocabulary covers heterogeneous domains.

Beyond within-super-category consistency, the dictionary also captures cross-super-category structure. Figure~\ref{fig:label_distance} visualizes pairwise cosine similarity between label prototypes (averaged dictionary entries per dominant part label), revealing three semantic patterns. \textbf{Within-object coherence:} strongest off-diagonal similarities link parts of the same super-category (\textit{Bird\_Body}--\textit{Bird\_Head}: $0.84$; \textit{Snake\_Body}--\textit{Snake\_Head}: $0.72$), reflecting shared appearance context despite spatial disjointness. \textbf{Taxonomic proximity:} reptiles and snakes form a coherent block (cosines $0.42$--$0.53$), and \textit{Bird\_Head}--\textit{Biped\_Head} reach $0.27$, capturing cross-category biological similarity without part-label supervision. \textbf{Functional analogy:} rigid parts cluster by geometric function (\textit{Car\_Tire}--\textit{Bike\_Tire}: $0.43$; \textit{Bottle\_Mouth}--\textit{Bottle\_Body}: $0.62$), while unrelated parts remain near zero. Together, these patterns show the dictionary discovers multi-level semantic organisation purely from the register bottleneck.

\begin{figure}[h]
\centering
\includegraphics[width=0.85\linewidth]{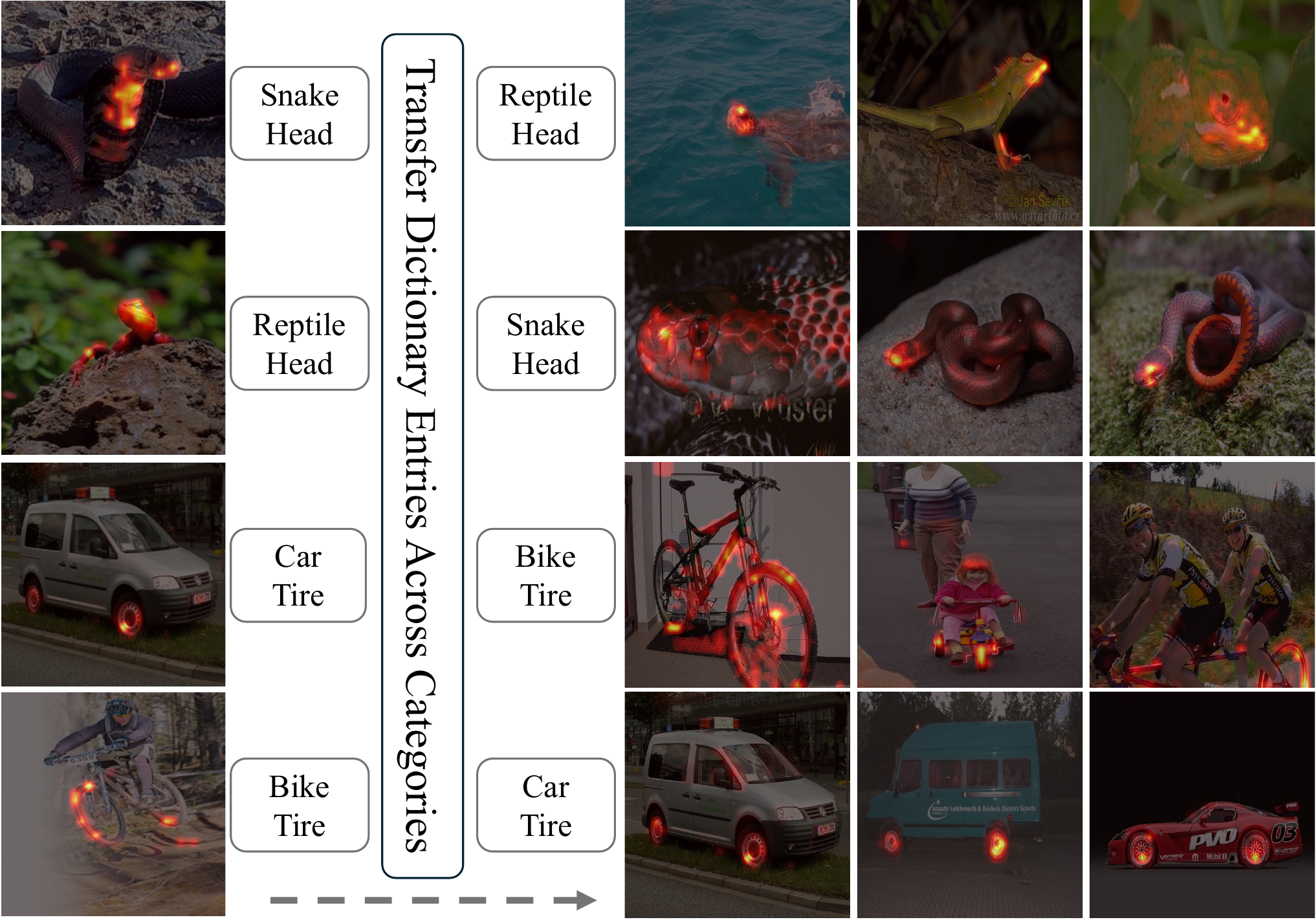}
\caption{\textbf{\textit{Cross super-category transfer of dictionary entries.}} The left column shows a source image whose registers activate a given entry (\textit{Snake\_Head}, \textit{Reptile\_Head}, \textit{Car\_Tire}, or \textit{Bike\_Tire}); the right three columns apply the \emph{same} entry to three images drawn from a \emph{different} super-category (labelled above the arrow). Rows~1--2 demonstrate taxonomic transfer between snake and reptile head entries (heatmap cosine $\approx 0.42$--$0.53$); rows~3--4 demonstrate functional transfer of the wheel concept between car and bike tyre entries (heatmap cosine $0.43$). The attention localises the analogous part in the target super-category.}
\label{fig:dict_cross_app}
\end{figure}

Figure~\ref{fig:dict_cross_app} qualitatively validates the latter two patterns. \textbf{Taxonomic transfer} (rows 1--2): \textit{Snake\_Head} and \textit{Reptile\_Head} entries transfer bidirectionally, localising heads across turtles, lizards, chameleons, and snakes despite large shape and texture differences. \textbf{Functional transfer} (rows 3--4): \textit{Car\_Tire} fires on bicycle wheels and \textit{Bike\_Tire} localises tyres on vans, buses, and race cars. These results confirm that high off-diagonal entries in Figure~\ref{fig:label_distance} reflect genuinely reusable part concepts rather than TF-IDF artefacts.

\subsection{Codebook construction and metrics}

\noindent\textbf{Dictionary construction.}
We compare two ways of building the codebook of $K$ entries: a
\emph{post-hoc K-means} dictionary built on PartImageNet, and a
\emph{learned} dictionary trained with the backbone on ImageNet.
A common labelling procedure, described at the end of this section,
attaches a part label to every entry after construction.

\smallskip
\noindent\emph{Post-hoc K-means dictionary.}
We extract register features $\{r_i\}$ from the last block of the trained
RATS model on every image of the PartImageNet training split. For each
register, we check whether its attention map peaks inside a ground-truth
ROI mask; only registers whose dominant attention falls within a
foreground part region are retained, filtering out registers that
primarily attend to background. We then run K-means on the pooled
foreground-filtered register features to obtain $K{=}512$ centroids
$\{e_k\}_{k=1}^{K}$ that serve directly as the dictionary entries.
PartImageNet contains $40$ fine-grained part categories ($11$
super-categories with $\sim\!4$ parts each) plus background; setting $K$
roughly an order of magnitude above this label count gives each part
several appearance-specific sub-prototypes (e.g.\ heads of different bird
species, tyres of different vehicle models).

\smallskip
\noindent\emph{Learned dictionary.}
We initialise $K$ entry vectors $\{e_k\}_{k=1}^{K}$ and optimise them
jointly with the RATS backbone on \textbf{ImageNet} under the standard
self-supervised objective. For each image, the registers
$\{r_i\}_{i=1}^{N}$ from the last block are hard-assigned to their
nearest entry, $k^{*}_{i}=\arg\min_{k}\|r_i-e_k\|_{2}$, and the entries
are pulled toward their assigned registers via a commitment\,+\,codebook
loss~\cite{oord2017neural}, with an entropy-based diversity term that prevents codebook
collapse. Because training sees ImageNet ($1000$ classes, covering parts
not annotated in PartImageNet) rather than PartImageNet's $\sim\!40$ part
labels, $K$ is scaled to the pre-training distribution: $K{=}4096$ as our
default, and $K{=}65{,}535$ for the large-vocabulary ablation.

\smallskip
\noindent\emph{Associating entries with semantic part labels.}
At this point the dictionary is just a set of $K$ unnamed vectors. To
turn it into a usable lexicon we need to attach a human-readable part
name (e.g.\ \emph{Bird\_Head}) to each entry. The intuition is simple:
\emph{when a register lands in entry $e_k$, what part of the image does
it tend to look at?} If almost every register that lands in $e_k$ looks
at bird heads, we call $e_k$ a \emph{Bird\_Head} entry. We make this
precise on the PartImageNet training split in three steps.

\smallskip
\noindent\textit{(i) Map each register to a dictionary entry.}
For every image, the $N$ registers from the last block are hard-assigned
to their nearest dictionary entry,
$k^{*}_{i}=\arg\min_{k}\|r_i-e_k\|_{2}$. This step uses no part
annotations; it is the same matching that the dictionary was built with.

\smallskip
\noindent\textit{(ii) Map each register to a ground-truth part.}
We now ask: which annotated part does this register attend to? Each
register has a spatial attention map over the patch grid (from the
L\,$\to$\,N stage of the RATS bottleneck), normalised to $[0,1]$. The
ground-truth segmentation provides a binary mask for every part in the
same image. We overlay the attention map onto each part mask and compute
the average attention value inside that mask. The part receiving the
highest average attention is taken as the register's part label.

\smallskip
\noindent\textit{(iii) Per-entry voting.}
After the previous two steps, every register carries two labels: which
dictionary entry it belongs to, and which ground-truth part it attends
to. For each entry, we count the part labels of all registers assigned
to it; the most frequent part becomes the entry's \emph{dominant part},
and its proportion is the entry's \emph{purity}. For example, if $100$
registers land in an entry and $82$ of them attend to
\emph{Bird\_Head}, the entry is named \emph{Bird\_Head} with a purity
of $0.82$. Higher purity indicates that the entry more consistently
corresponds to a single part. Entries receiving fewer than three
registers are discarded, as too few samples cannot yield a reliable
statistic.

\smallskip
\noindent\emph{Inference with the dictionary.}
Given a new image, we run a single forward pass through the RATS model to
obtain each register's feature $r_i$ and attention map $A_i$. Matching
$r_i$ to its nearest dictionary entry assigns the register the
corresponding part label, and the attention map indicates where that part
lies in the image. Since multiple registers may be assigned the same part
(e.g.\ several registers all match to \emph{Bird\_Head}), we merge their
attention maps to produce a heatmap that covers the full extent of that
part in the image (see Fig.~\ref{fig:dict_qualitative} and
Fig.~\ref{fig:codebook_ood}).

\noindent\textbf{Bag-of-entries representation.}
Given the register-to-entry assignment described above, we can describe
an image by which dictionary entries its registers fall into and how many
land in each~\cite{sivic2003video}, forming a $K$-dimensional count vector
$\mathrm{tf}_{j} \in \mathbb{R}^{K}$. Using raw counts directly would
bias toward entries that appear frequently across all images (e.g.\
background). We therefore borrow the TF--IDF scheme~\cite{salton1988term} from information
retrieval: the more images an entry appears in, the lower its weight
($\mathrm{idf}_k = \log \frac{|\mathcal{D}|}{|\{j : \mathrm{tf}_{j,k} > 0\}|}$,
where $|\mathcal{D}|$ is the total number of images). After weighting and
$\ell_2$ normalisation, the final representation is
$\mathbf{v}_j = (\mathrm{tf}_j \odot \mathrm{idf})\,/\,
\|\mathrm{tf}_j \odot \mathrm{idf}\|_2$.
Image retrieval then ranks database images by cosine similarity against
the query.

\noindent\textbf{Metrics.}
\emph{mAP} (mean Average Precision) measures whether the dictionary can
help retrieve similar images. Concretely, we take each image, use its
bag-of-entries vector to find the most similar images in the database,
and check whether the retrieved images belong to the same object category
as the query (e.g.\ both are birds, both are fish). Repeating this for
all images and averaging gives mAP. \emph{R@1} is a simplified variant
that only checks whether the single most similar image shares the
query's category; it saturates quickly on this benchmark and offers
limited discriminative power.

\emph{valPw} measures how specialised the dictionary entries are.
Ideally, each entry should correspond to exactly one part (e.g.\ only
bird heads). valPw is the weighted average of all entries' purities,
where the weight of each entry is the number of registers assigned to
it---entries matched by more registers are more important and thus count
more in the average. Only entries matched by at least three registers
are included.

\emph{util\%} measures how many of the $K$ entries are actually used
(i.e.\ matched by at least three registers). For example, if
$K{=}4096$ but only $400$ entries are used, util\% is roughly $10\%$,
indicating that most of the dictionary capacity is wasted.

\section{Per-head register attention maps}
\label{app:perhead_attn}

\begin{figure}[h]
\centering
\includegraphics[width=\linewidth]{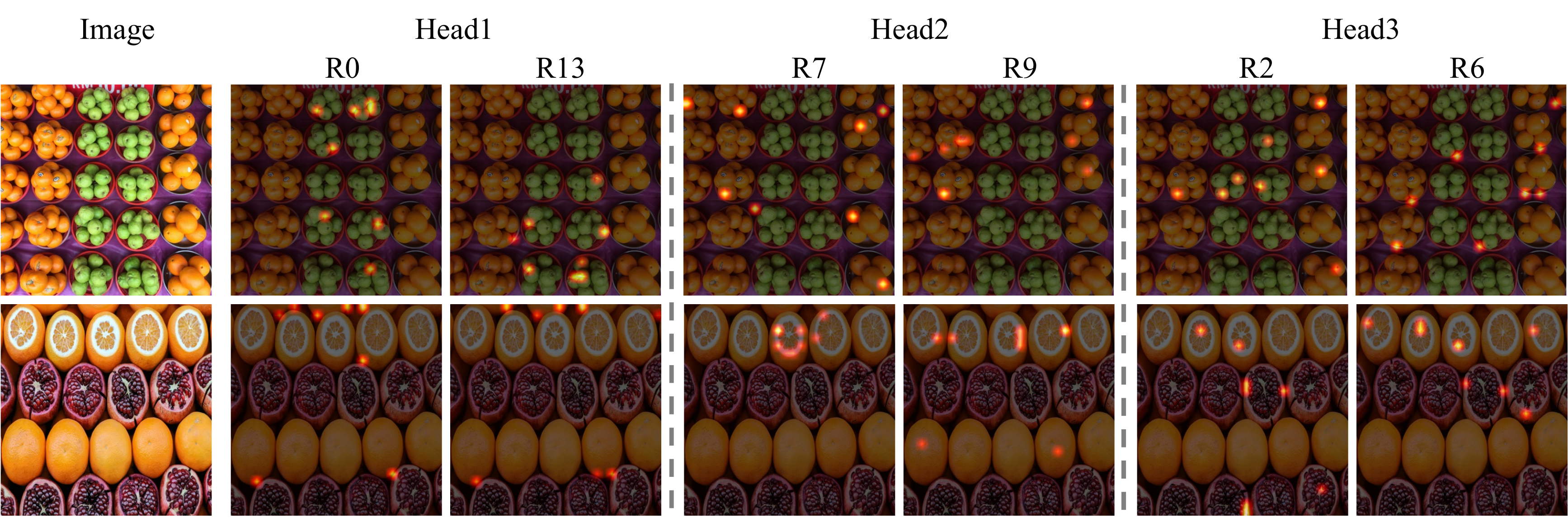}
\caption{\textbf{\textit{Per-head register attention maps.}} Six selected registers from three heads on two dense-instance images; analysed in Sec.~\ref{app:perhead_attn}.}
\label{fig:attn_appendix}
\end{figure}

Figure~\ref{fig:attn_appendix} shows attention maps from three heads of
the last block on two images containing dense object instances
(a fruit grid in row~1 and a sliced-fruit row in row~2).
For each head we display two registers and visualise their $L{\to}N$
attention overlaid on the image.
Two qualitative observations stand out.
\textbf{(i)}~Within a single head, distinct registers focus on different
but often related spatial regions --- e.g., $R_0$ and $R_{13}$ in
Head~1 both fire on green-fruit clusters but at different positions.
\textbf{(ii)}~Across heads, the same image region is covered by
different register combinations: Head~1 and Head~2 emphasise different
inter-fruit boundaries, while Head~3 captures cross-row alignment.
These patterns are consistent with the per-head register separation
acting as an inductive bias toward diverse spatial decompositions of
the input.

\begin{figure}[!h]
\centering
\includegraphics[width=\linewidth]{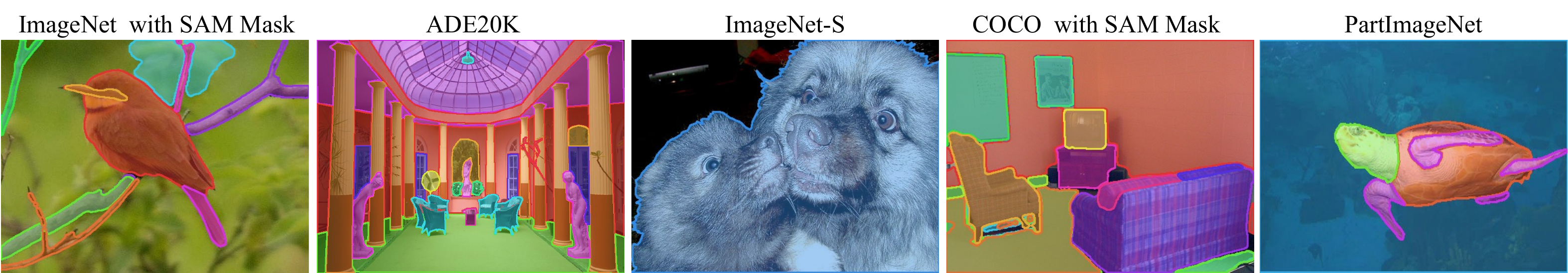}
\caption{\textbf{Ground-truth annotations across the five evaluation benchmarks.} ImageNet and COCO use SAM~2.1 generated masks; ADE20K, ImageNet-S$_{919}$, and PartImageNet use their native annotations.}
\label{fig:datasets}
\end{figure}

\section{Implementation details.}
\label{app:impl}

\paragraph{Architecture.}
We provide two model scales: RATS-S and RATS-B, based on ViT-S/16 ($D{=}384$, $H{=}6$) and ViT-B/16 ($D{=}768$, $H{=}12$), both with 12 layers. In each block, MHSA is replaced by the region bottleneck attention. Registers receive no positional encoding and are hard-partitioned into $n{=}N/H$ per head. By default, RATS-S uses $N{=}96$ ($n{=}16$) and RATS-B uses $N{=}192$ ($n{=}16$).

\paragraph{Pretraining.}
Both models are pretrained on ImageNet-1k at $224{\times}224$, following the DINO recipe: AdamW~\cite{loshchilov2019decoupled} optimizer, cosine weight decay from 0.04 to 0.4, linear warmup (10 epochs) followed by cosine learning rate decay~\cite{loshchilov2017sgdr}, teacher EMA momentum $\lambda{=}0.996{\to}1.0$, output dimension $K{=}65{,}536$, two global crops ($224^2$) plus ten local crops ($96^2$), teacher temperature $\tau_t{=}0.07$, and bf16 mixed precision~\cite{micikevicius2018mixed} throughout. RATS-B fully aligns with the official DINO ViT-B hyperparameters: 400 epochs, total batch size 1024, base learning rate $7.5{\times}10^{-4}$, minimum learning rate $2{\times}10^{-6}$, drop path~\cite{huang2016deep} 0.1, gradient clipping 0.3, teacher temperature warmup for 50 epochs, and freeze last layer for 3 epochs. RATS-S follows: 300 epochs, total batch size 768 (96 per GPU $\times$ 8 GPUs), base learning rate $2.5{\times}10^{-4}$, minimum learning rate $1{\times}10^{-5}$, drop path 0.05, gradient clipping disabled, teacher temperature warmup for 30 epochs, and freeze last layer for 1 epoch. The training objective is solely the DINO student--teacher cross-entropy.

\paragraph{Datasets and evaluation.}
Evaluating part discovery requires dense region annotations that cover the entire image. However, ImageNet lacks segmentation masks entirely, and COCO, while providing instance annotations, has too few masks with limited spatial coverage to adequately evaluate full-image region grouping. We therefore adopt SAM~2.1 as proxy ground truth for these two datasets: we run the official \texttt{sam2.1-hiera-large} automatic mask generator with a $128{\times}128$ point prompt grid to produce dense, category-agnostic region maps on each image. The remaining three datasets use their native annotations.

We evaluate on five benchmarks: COCO 2017 val (5,000 images, SAM masks), testing grouping in cluttered multi-object scenes; ADE20K validation (2,000 images), testing robustness under large semantic diversity; ImageNet val (50,000 images, SAM masks), testing part-level grouping on object-centric images; ImageNet-S$_{919}$ val (10,767 images), biased toward foreground and part evaluation; and PartImageNet test (2,408 images), with human part annotations. Figure~\ref{fig:datasets} shows representative examples with their ground-truth annotations from each benchmark.

\paragraph{Metrics.}
We report many-to-one mIoU (M2O) as the primary metric, which greedily assigns each predicted region to the ground-truth region with maximum overlap, allowing multiple predictions to map to the same ground truth. We additionally report ARI (Adjusted Rand Index)~\cite{hubert1985comparing}, which measures agreement between predicted and ground-truth groupings without relying on region-count matching. In ablation studies, we further report one-to-one mIoU (O2O, via Hungarian matching~\cite{kuhn1955hungarian}): a large M2O--O2O gap indicates that the model fragments a single semantic region across multiple registers. All segmentation methods are evaluated at $512{\times}512$. Higher resolution provides finer patch grids for the register bottleneck while remaining computationally feasible for all baselines. We ablate resolution sensitivity in Figure~\ref{fig:resolution}.

\paragraph{Downstream transfer setting.}
We fine-tune the pretrained RATS and ViT-S/16 backbone with a Mask2Former~\cite{cheng2022masked} decoder on two benchmarks. For ADE20K semantic segmentation, we follow the standard MMSegmentation~\cite{mmseg2020} recipe (crop $512{\times}512$, 160K iterations, batch size 8, 96 object queries). For COCO 2017 detection and instance segmentation, we follow the standard MMDetection~\cite{mmdet2019} recipe ($1024{\times}1024$ input, 368.75K iterations, batch size 8, 96 object queries, AdamW). The baseline uses the same Mask2Former architecture initialized from a DINO~\cite{caron2021emerging} ViT-S/16 checkpoint trained for the same 100 epochs, with all fine-tuning hyperparameters kept identical.

During fine-tuning, we restore a parallel $L{\to}L$ self-attention path alongside the register bottleneck, so that each block computes
\begin{equation}
  X_{\text{out}} = \underbrace{(L{\to}N{\to}N{\to}L)(X)}_{\text{bottleneck}} + \underbrace{(L{\to}L)(X)}_{\text{full-rank}}.
\end{equation}
The bottleneck path extracts part-level representations that serve as initial queries for the Mask2Former decoder. However, this path alone is low-rank: each head operates through only $n{=}N/H$ registers (e.g., $n{=}16$), producing an attention matrix of rank at most $n$, which is insufficient to maintain the full expressive capacity of a ViT whose representation rank scales with the patch sequence length $L$. The $L{\to}L$ path restores standard self-attention among all patch tokens, preserving full-rank capacity and ensuring that the fine-tuned model retains the dense representational power required for high-resolution segmentation and detection.